\documentclass[runningheads]{llncs}

\usepackage[T1]{fontenc}
\usepackage{amsmath,amssymb,amsfonts}
\usepackage{algorithm2e}
\usepackage{bm}
\usepackage{graphicx}
\usepackage{textcomp}
\usepackage{xcolor}
\usepackage{booktabs}
\usepackage{multirow}
\usepackage{placeins}
\usepackage{listings}
\usepackage{enumitem}
\usepackage{bbm}
\usepackage{hyperref}

\newcommand{\ind}[1]{\mathbbm{1}\{#1\}}

\newcommand{\TextEncName}{ModernBERT}        
\newcommand{\TextEncAlt}{DeBERTa-v3}
\newcommand{\VisEncName}{SigLIP-2}
\newcommand{\simSig}{\mathrm{sim}_{\text{\VisEncName}}} 
\newcommand{\tprojEnt}{P_t^{\mathrm{ent}}}
\newcommand{\tprojPair}{P_t^{\mathrm{pair}}}
\newcommand{\vproj}{P_v}

\newcommand{\Fuse}{\operatorname{Fuse}}
\newcommand{\SpanRep}{\operatorname{SpanRep}}
\newcommand{\TopK}{\operatorname{TopK}}
\newcommand{\up}{\ensuremath{\uparrow}}
\newcommand{\down}{\ensuremath{\downarrow}}

\begin{document}

\title{SAVER: Selective As-Needed Vision Evidence for Multimodal Information Extraction}
\titlerunning{SAVER: Selective Vision Evidence for Multimodal IE}

\newcommand{\orcidAuthorOne}{\href{https://orcid.org/0000-0002-6417-3280}{\protect\includegraphics[scale=0.045]{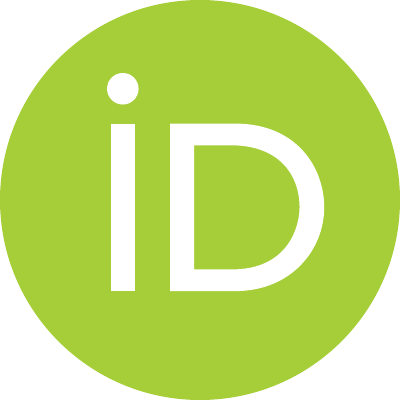}}}
\newcommand{\orcidAuthorTwo}{\href{https://orcid.org/0009-0007-5519-6222}{\protect\includegraphics[scale=0.045]{pic/orcid.pdf}}}
\newcommand{\orcidAuthorThree}{\href{https://orcid.org/0009-0008-0501-8652}{\protect\includegraphics[scale=0.045]{pic/orcid.pdf}}}
\newcommand{\orcidAuthorFour}{\href{https://orcid.org/0009-0000-6038-7831} 
{\protect\includegraphics[scale=0.045]{pic/orcid.pdf}}}
\newcommand{\orcidAuthorFive}{\href{https://orcid.org/0000-0002-1069-4830}{\protect\includegraphics[scale=0.045]{pic/orcid.pdf}}}
\newcommand{\orcidAuthorSix}{\href{https://orcid.org/0009-0007-2672-6650}{\protect\includegraphics[scale=0.045]{pic/orcid.pdf}}}
\newcommand{\orcidAuthorSeven}{\href{https://orcid.org/0009-0002-6042-3454}{\protect\includegraphics[scale=0.045]{pic/orcid.pdf}}}
\newcommand{\orcidAuthorEight}{\href{https://orcid.org/0000-0002-1799-3948}{\protect\includegraphics[scale=0.045]{pic/orcid.pdf}}}

\author{
Miaobo Hu\inst{1,2}\orcidAuthorOne \and
Bokun Wang\inst{1,2}\orcidAuthorTwo \and
Shuhao Hu\inst{1,2}\orcidAuthorThree \and
Rui Chen\inst{1,2}\orcidAuthorFour \and
Xin Wang\inst{1}\orcidAuthorFive \and
Xiaobo Guo\inst{1}\orcidAuthorSix \thanks{Corresponding author: guoxiaobo@iie.ac.cn} \and
Daren Zha\inst{1}\orcidAuthorSeven \and
Jun Xiao\inst{3}\orcidAuthorEight
}

\authorrunning{M. Hu et al.}

\institute{
Institute of Information Engineering, Chinese Academy of Sciences, Beijing 100092, China
\email{\{humiaobo,wangbokun,hushuhao,chenrui2025\}@iie.ac.cn}
\email{\{wangxin,guoxiaobo,zhadaren\}@iie.ac.cn}
\and
School of Cyber Security, University of Chinese Academy of Sciences, Beijing 100049, China
\and
School of Artificial Intelligence, University of Chinese Academy of Sciences, Beijing 100049, China
\email{xiaojun@ucas.ac.cn}
}



\maketitle

\begin{abstract}
Multimodal IE in social media is difficult because a post may attach multiple images that are weakly related, redundant, or even misleading with respect to the text. In this setting, always-on multimodal fusion wastes computation and can amplify spurious visual cues. The core challenge is to decide, for each candidate span or marked entity pair, whether vision should be consulted at all and, if so, which small subset of images provides trustworthy evidence.

We propose \textbf{SAVER}, a selective vision-as-needed framework for multimodal named entity recognition and multimodal relation extraction. SAVER uses a \emph{Conformal Groundability Gate (CGG)} to estimate span-level visual groundability in MNER, derive pair-level activation in MRE from the two marked entities, and calibrate the activation threshold on a held-out split via a conformal-style procedure with Clopper--Pearson upper bounds. When activated, a submodular relevance--diversity selector chooses a compact evidence subset across images, which is then aggregated by a Set Transformer. An energy-inspired joint scoring head combines text, optional visual evidence, text--image consistency, and sparse routing for entity typing or relation classification.

Experiments show that SAVER consistently improves F1 over strong text-only and always-on multimodal baselines, while reducing AURC, increasing activation coverage at a fixed risk level, and lowering FLOPs and P90 latency.

\keywords{Multimodal Information Extraction \and Selective Prediction \and Conformal-Style Calibration \and Submodular Selection \and Multi-image Evidence}
\end{abstract}

\section{Introduction}
\label{sec:intro}

Multimodal information extraction (IE) on social media is challenging because posts often attach multiple images that are only loosely related to the text. Some images are useful, some are redundant, and some are misleading. This changes the core problem from ``how to fuse text and images'' to \emph{when vision should be used at all, and which small subset of images should be trusted for the current prediction unit}. For MNER, the unit is a candidate entity span; for MRE, it is the marked entity pair.

This problem is difficult for three reasons. First, training data typically annotate entities and relations but do not annotate whether a unit is visually groundable. Second, image relevance varies substantially across posts, so always-on fusion can amplify spurious cross-modal correlations. Third, in multi-image settings, accuracy and efficiency are coupled: using all images increases compute and latency, while aggressive pruning risks discarding useful evidence.

We address this with \textbf{SAVER} (\emph{Selective As-Needed Vision Evidence}), which treats vision as \emph{optional evidence} rather than a mandatory input channel. SAVER uses a \emph{Conformal Groundability Gate (CGG)} to estimate whether a unit is likely to benefit from visual evidence and to calibrate the activation threshold under a target activated-subset risk. When vision is activated, a submodular selector (SIS; with an optional RL-based selector, RES, described in Sec.~\ref{sec:rl}) constructs a small, diverse evidence set, which is aggregated by a Set Transformer. An energy-inspired joint scoring head then combines text, optional visual evidence, consistency regularization, and sparse routing for entity typing or relation classification. Because CGG relies only on cached global image embeddings and SIS optimizes relevance plus diversity, the framework is both computationally feasible and interpretable.

\noindent\textbf{Contributions.}
\begin{itemize}[leftmargin=*]
\item \textbf{Risk-controlled gating for vision.} CGG calibrates an activation threshold on a held-out split via a conformal-style selection rule with Clopper--Pearson upper bounds, targeting a bounded activated-subset error.
\item \textbf{Compact multi-image evidence.} SIS (and an optional RL-based selector, RES) selects a small, diverse evidence subset and aggregates it permutation-invariantly with a Set Transformer, reducing redundancy and cost.
\item \textbf{Unified scoring with consistency \& sparsity.} An energy-inspired joint scoring head couples MNER span/type prediction and MRE relation classification with text--image consistency and sparse vision routing, improving accuracy and selectivity while preserving the efficiency gains from selective routing across both single- and multi-image benchmarks.
\end{itemize}

\section{Related Work}
\label{sec:relwork}

\noindent\textbf{Multimodal IE and multi-image settings.}
Early multimodal IE typically treats images as always-on enhancement signals by directly concatenating or attentively fusing visual and textual features, which can amplify spurious cues when images are irrelevant. Representative examples include HVPNeT, which injects hierarchical visual prefixes \cite{chen_good_2022}, MoRe-style retrieval augmentation \cite{wang_named_2022}, and RSRNeT, which unifies MNER and MRE but still uses visual information in an always-on manner \cite{wang_rsrnet_2024}. On the data side, MNRE established a standard single-image RE benchmark \cite{zheng_mnre_2021}, while MRE-MI, MNER-MI, and MNER-MI-Plus move closer to real social-media posts with multiple attached images \cite{huang_mner-mi_2024,huang_mre-mi_2025}. These datasets motivate explicit evidence selection rather than uniform attention over all images.

\noindent\textbf{Selective prediction and conformal gating.}
Selective prediction studies the trade-off between coverage and risk and supports abstention or routing under user-specified error levels \cite{el-yaniv_foundations_2010,geifman_selective_2017,geifman_selectivenet_2019}. Conformal risk control provides finite-sample upper bounds for target risk under calibration-time i.i.d.\ assumptions \cite{angelopoulos_conformal_2025}. Our CGG adapts this line of work to multimodal IE by calibrating span-level visual groundability scores in MNER together with a derived pair-level routing rule for MRE, targeting bounded activated-subset error.

\noindent\textbf{Evidence selection, set aggregation, and structured decoding.}
When multiple images or regions are available, evidence selection requires a relevance--diversity trade-off. Submodular maximization and determinantal point processes are standard diversity-aware choices \cite{bordeaux_submodular_2014,kulesza_determinantal_2012}, and Set Transformer provides a natural permutation-invariant aggregator for compact evidence sets \cite{lee_set_2019}. Structured energy networks further motivate coupling span typing and relation prediction rather than optimizing them in isolation \cite{belanger_structured_2016}. SAVER combines these ideas through risk-controlled gating, submodular or RL-based evidence acquisition, permutation-invariant aggregation, and an energy-inspired joint scoring objective.

\section{Method}
\label{sec:method}

SAVER follows a simple principle: use vision only when the current entity or marked entity pair is likely to be visually grounded, and when activated, acquire only a small complementary evidence set. This yields a modular pipeline with risk-bounded activation, compact multi-image evidence construction, and unified downstream scoring.

\subsection{Task setting and notation}
A social media sample consists of text $x=(w_1,\dots,w_T)$ and images $\mathcal{I}=\{I_1,\dots,I_N\}$. SAVER is instantiated in two settings.

\noindent\textbf{MNER.} We enumerate candidate spans $s$ in the text. Let $\delta_s\in\{0,1\}$ denote the span-validity variable for candidate span $s$, and let $y_s$ denote its entity type when $\delta_s=1$. In the gold annotation, $\delta_s=1$ iff $s$ is an entity span. The output can be written as
\begin{align}
Y^{\mathrm{ner}}=(\mathcal{S},\mathcal{T}),
\end{align}
where $\mathcal{S}=\{s:\delta_s=1\}$ and $\mathcal{T}=\{y_s\}_{s\in\mathcal{S}}$.

\noindent\textbf{MRE.} Following MNRE and MRE-MI, the input contains a marked entity pair represented by head and tail spans $(s_h,s_t)$ in the text, and the target is the relation label
\begin{align}
Y^{\mathrm{re}}=r_{(h,t)}.
\end{align}
For brevity, pair-level quantities associated with the marked pair $(s_h,s_t)$ are indexed by $(h,t)$.

We use a text encoder to obtain token representations $H\in\mathbb{R}^{T\times d}$. A vision encoder produces one global vector per image, while region-level vectors are extracted only for selected images when the gate is active. We denote the default submodular main model by \emph{SAVER-SIS}, which is the primary model reported in the tables. The optional RL-based selector is called \emph{Reinforced Evidence Selector (RES)}, and the corresponding model variant is \emph{SAVER-RL}.

\subsection{Overall pipeline}
Given $(x,\mathcal{I})$, we perform four steps:
\begin{enumerate}[leftmargin=*]
\item \textbf{Encoding:} The text encoder produces token/span representations; the vision encoder produces a global vector per image, while region vectors are extracted only for images selected after gating.
\item \textbf{CGG (unit-level whether-to-use-vision):} For each candidate entity span $s$ in MNER, compute a groundability score; for MRE, derive a pair-level gate for the marked entity pair $(s_h,s_t)$ from the two entity scores, and trigger the visual branch under a calibrated threshold.
\item \textbf{Evidence Constructor (SIS or RES) + Set Transformer:} If the relevant gate is active, either (i) maximize a submodular objective to select $K$ images, or (ii) run the optional \emph{Reinforced Evidence Selector (RES)}, an RL policy that sequentially acquires evidence with a STOP action under cost-aware rewards and CGG-based action masking; the selected evidence is then aggregated by a \emph{Set Transformer}.
\item \textbf{Energy-Inspired Joint Scoring:} Inject text and optional visual evidence into a unified scoring function. For MNER, we score spans and types; for MRE, we score the relation of the marked entity pair.
\end{enumerate}

\begin{figure}[tb] 
  \centering
  \includegraphics[width=\linewidth]{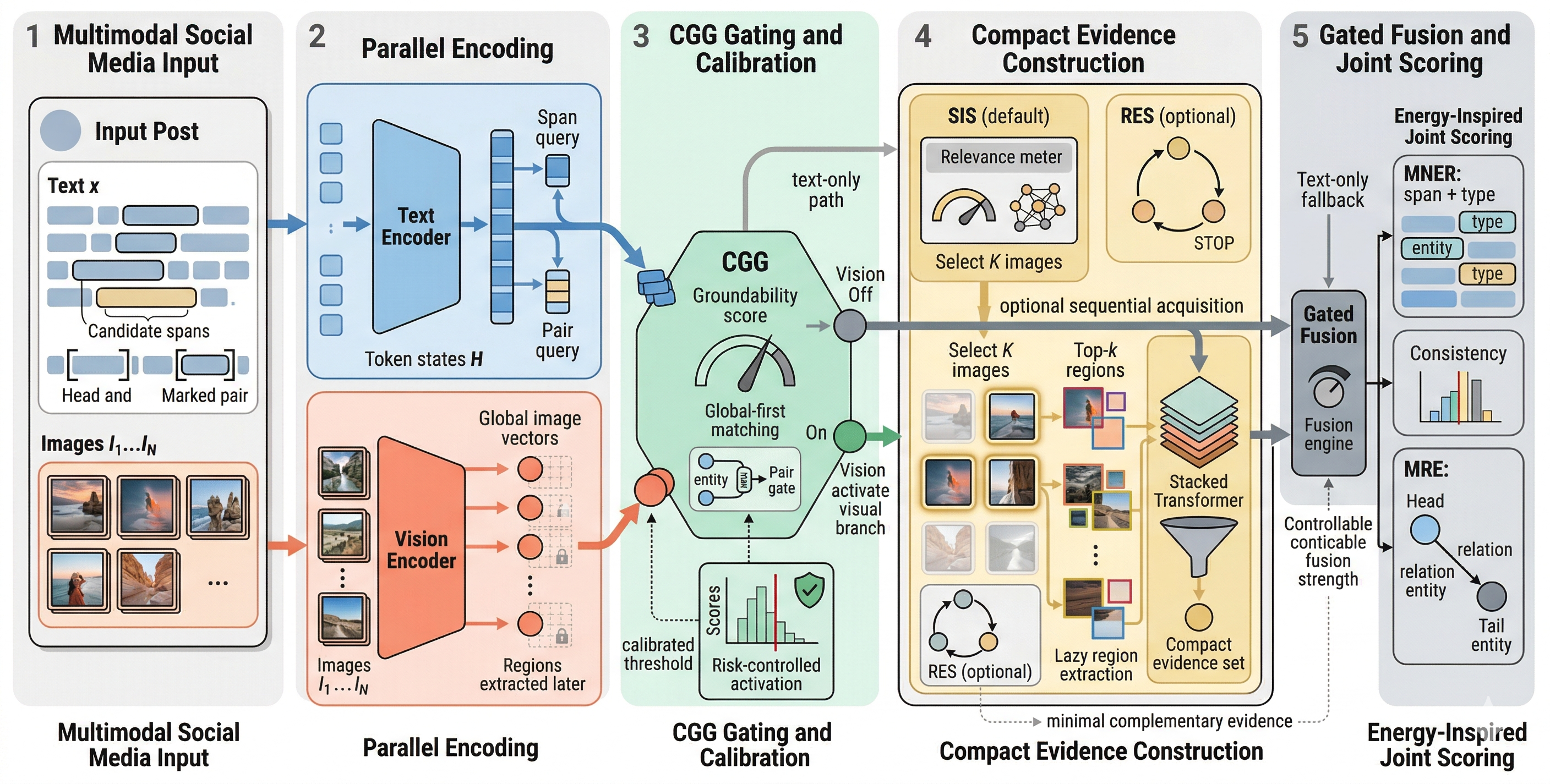}
  \caption{SAVER overview: text and vision are encoded in parallel; CGG decides whether to activate vision; when activated, SIS/RES together with a Set Transformer build a compact multi-image evidence set that is fused with text before energy-inspired joint scoring for MNER or MRE.}
  \label{fig:saver_overview}
\end{figure}

\subsection{Representations and Fusion}
\label{subsec:enc}

The text encoder yields token states $H=[h_1,\dots,h_T]\in\mathbb{R}^{T\times d}$. For an entity span $s=(a,b)$, we use
\begin{align}
h_s=\SpanRep(H,a,b),\qquad q_s=\tprojEnt(h_s)\in\mathbb{R}^{d_c}.
\end{align}

For MRE, let $s_h$ and $s_t$ denote the marked head and tail spans, with span representations $h_{s_h}$ and $h_{s_t}$. We compose
\begin{align}
u_{(h,t)}=[h_{s_h};h_{s_t};\phi_{\text{dist}}(s_h,s_t)],\qquad
q_{(h,t)}=\tprojPair(u_{(h,t)}).
\end{align}
Here $\phi_{\text{dist}}(\cdot)$ encodes relative distance and order features between the head and tail spans.
SAB and PMA denote the standard Set Transformer self-attention block and pooling by multihead attention, respectively.
We use $f_{\text{cons}}(x)=-x$ to reward higher text--image consistency.

For each image $I_i$, the vision encoder outputs a global vector $z_i$ and region vectors $\{z_{i,m}\}$, projected as
\begin{align}
v_i=\vproj(z_i),\qquad v_{i,m}=\vproj(z_{i,m}),
\end{align}
with cosine similarity
\begin{align}
\simSig(a,b)=\frac{\langle a,b\rangle}{\|a\|\,\|b\|}.
\end{align}

If CGG activates vision for a query unit $u$ (an entity in MNER or the marked pair in MRE), SIS/RES selects an image subset $A\subseteq\{1,\dots,N\}$, $|A|\le K$. Let $q_u$ denote the corresponding query embedding. For each selected image $i\in A$, let $\TopK(i)$ denote the indices of the top-$k$ regions ranked by $\simSig(q_u,v_{i,m})$. We then form
\begin{align}
\mathcal V_A=\{v_i\}_{i\in A}\cup\{v_{i,m}\}_{i\in A,\,m\in \TopK(i)}.
\end{align}
A Set Transformer aggregates the evidence:
\begin{align}
V'=\operatorname{SAB}(\mathcal V_A),\qquad z_u^{\text{set}}=\operatorname{PMA}(V').
\end{align}

For an entity span $s$, gated fusion is
\begin{align}
\tilde h_s=(1-\eta_s)h_s+\eta_s\,\Fuse(h_s,z_s^{\text{set}}),\qquad
\eta_s=
\begin{cases}
g(s), & \text{train},\\
g(s)\gamma(s), & \text{test},
\end{cases}
\end{align}
and the consistency score is
\begin{align}
\mathrm{Consis}(s)=\simSig(\tprojEnt(\tilde h_s), z_s^{\text{set}}).
\end{align}
This quantity is evaluated only when visual evidence is constructed; otherwise we set $\mathrm{Consis}(s)=0$ and use the text-only representation $\tilde h_s=h_s$.

For MRE, we use a simple low-cost pair gate derived from the two marked entities:
\begin{align}
g_{(h,t)}=\max(g(s_h),g(s_t)),\qquad
\gamma_{(h,t)}=\ind{g_{(h,t)}\ge \tau}.
\end{align}
This rule activates pair-level vision whenever either endpoint appears visually groundable, avoiding an additional pair-specific gating network.
If pair-level evidence is built, we use
\begin{align}
\tilde u_{(h,t)}=
(1-\eta_{(h,t)})u_{(h,t)}+\eta_{(h,t)}\,\Fuse(u_{(h,t)},z_{(h,t)}^{\text{set}}),
\end{align}
with
\begin{align}
\eta_{(h,t)}=
\begin{cases}
g_{(h,t)}, & \text{train},\\
g_{(h,t)}\gamma_{(h,t)}, & \text{test},
\end{cases}
\end{align}
and pair-level consistency
\begin{align}
\mathrm{Consis}(h,t)=\simSig\!\big(\tprojPair(\tilde u_{(h,t)}), z_{(h,t)}^{\text{set}}\big).
\end{align}
If no pair-level evidence is constructed, we fall back to the text-only pair representation
\begin{align}
\tilde u_{(h,t)}=[h_{s_h};h_{s_t};\phi_{\text{dist}}(s_h,s_t)],
\end{align}
and set
\begin{align}
\mathrm{Consis}(h,t)=0.
\end{align}
During training, we materialize evidence for supervised units and use soft gates $\eta$ for differentiable weighting; during inference, evidence is constructed only for hard-activated units with $\gamma(\cdot)=1$.

\subsection{Conformal Groundability Gate (CGG)}
\label{subsec:cgg}
\subsubsection{Intuition}
We use \emph{visual groundability} to denote whether a text unit can be reliably matched to visual evidence. As the need detector, CGG assigns each candidate span a groundability score $g(s)$ and, after calibration, activates the visual branch only when this score is sufficiently high; otherwise, the model follows the \emph{text-only} path to avoid noise and extra cost. Empirically, CGG tends to assign higher scores to concrete visual entities such as faces, logos, landmarks, or objects, and lower scores to abstract sentiments or discourse markers, even though no explicit groundability labels are provided.

\subsubsection{Groundability score (global-first, lazy region extraction)}
For entity $s$, we compute $g(s)$ using \emph{only global image vectors} $\{v_i\}_{i=1}^N$; region-level vectors $\{v_{i,m}\}$ are \emph{materialized lazily} \textbf{only} for images selected downstream (SIS/RES), avoiding unnecessary region encodings when the gate is off:
\begin{align}
g(s) &= \sigma\!\Big(W_g\,[\,h_s;\ \psi_{\max}^{\text{glob}}(q_s,\{v_i\});\ \psi_{\text{stat}}^{\text{glob}}(q_s,\{v_i\})\,]\Big)
\end{align}
\begin{align}
\psi_{\max}^{\text{glob}}(q_s,\{v_i\}) &= \max_{1\le i\le N}\ \simSig\!\big(q_s, v_i\big).
\end{align}
Here $\{v_i\}$ are computed once per image (cacheable); in experiments, 
$\psi_{\text{stat}}^{\text{glob}}(q_s,\{v_i\})$ concatenates summary statistics from 
$\{\simSig(q_s,v_i)\}_{i=1}^N$, namely [mean, std, top-2 mean], where the top-2 mean averages the largest $\min(2,N)$ similarities; the maximum is handled separately by $\psi_{\max}^{\text{glob}}$. 
Region-level $\{v_{i,m}\}$ are extracted \emph{after} gating and subset selection, yielding latency/FLOPs savings when $\gamma(\cdot)=0$. 

\subsubsection{Gating decision and split-calibrated threshold selection}
\label{sec:cgg-crc}

For notational convenience, let
\begin{equation}
g(u)=
\begin{cases}
g(s), & u=s \text{ (MNER)},\\
g_{(h,t)}, & u=(s_h,s_t) \text{ (MRE)}.
\end{cases}
\end{equation}
We activate the visual branch when
\begin{equation}
\gamma(u)=\ind{g(u)\ge \tau}.
\end{equation}

We split off a held-out calibration set $\mathcal{C}$. 
For each calibration unit $u\in\mathcal{C}$, we run the decoder with the gate forced on for $u$ and record the binary downstream loss
\begin{equation}
\ell(u)=
\begin{cases}
\ind\{\hat\delta_s\neq\delta_s \lor (\delta_s=1 \land \hat y_s\neq y_s)\}, & u=s \text{ (MNER)},\\
\ind\{\hat r_{(h,t)}\neq r_{(h,t)}\}, & u=(s_h,s_t) \text{ (MRE)}.
\end{cases}
\end{equation}

For a candidate threshold $\tau'$, define the activated calibration subset
\begin{align}
\mathcal{C}_{\tau'}=\{u\in\mathcal{C}: g(u)\ge \tau'\},
\end{align}
with size $n(\tau')=|\mathcal{C}_{\tau'}|$ and number of errors
\begin{align}
k(\tau')=\sum_{u\in\mathcal{C}_{\tau'}} \ell(u).
\end{align}
We then select the threshold with the largest activation coverage whose Clopper--Pearson upper bound on activated-subset error is below a user-chosen target risk $\alpha$:
\begin{equation}
\tau
=
\arg\max_{\tau'}
\frac{n(\tau')}{|\mathcal{C}|}
\quad
\text{s.t.}
\quad
\operatorname{CPUpper}\bigl(k(\tau'), n(\tau'); 1-\delta\bigr)\le \alpha.
\end{equation}
Here $1-\delta$ is the desired confidence level, and $\operatorname{CPUpper}(k,n;1-\delta)$ denotes the exact Clopper--Pearson upper confidence bound for a binomial proportion.
This is a split-calibrated, conformal-style threshold-selection rule, where ``conformal-style'' denotes split calibration with binomial upper confidence bounds rather than full set-valued conformal prediction. CGG is trained end-to-end through the downstream task loss together with the sparsity regularizer $\lambda_{\text{gate}}\sum_u \eta_u$, without explicit groundability supervision. Under the usual i.i.d.\ calibration/test assumption, it yields a calibrated operating point for vision activation under the selected Clopper--Pearson criterion; under distribution shift, we re-calibrate $\tau$ on a small held-out slice from the target domain.

\subsection{Evidence Constructor: SIS or RES}
\label{subsec:evidence}

When vision is needed, SAVER constructs a minimal yet informative evidence set. Our main model uses the \emph{Submodular Image Selector (SIS)}; we also describe an optional RL extension, \emph{Reinforced Evidence Selector (RES)}, which performs sequential acquisition with a STOP action.

\subsubsection{SIS: monotone submodular selection}
\label{sec:sis_obj}

If the relevant gate is active, we select an image subset $A\subseteq \{1,\dots,N\}$ with $|A|\le K$. We use $\gamma(s)$ for entity-level selection and $\gamma_{(h,t)}$ for pair-level selection. Let $q$ denote the projected query embedding, namely $q=q_s$ for entity-level selection and $q=q_{(h,t)}$ for pair-level selection. We then define
\begin{align}
r_i &= \simSig(q,v_i),\\
d_{ij} &= \simSig(v_i,v_j).
\end{align}
To keep the facility-location objective non-negative, we rescale cosine similarities into $[0,1]$:
\begin{align}
\tilde r_i &= \frac{1+r_i}{2},\qquad
\tilde d_{ij} = \frac{1+d_{ij}}{2}. \label{eq:sis_rescale}
\end{align}
We maximize the monotone submodular objective
\begin{align}
F_{\text{SIS}}(A)
&=
\lambda_{\text{rel}}\sum_{i\in A}\tilde r_i
+
\lambda_{\text{cov}}\sum_{j=1}^{N}\tilde r_j \max_{i\in A}\tilde d_{ij},
\label{eq:sis_obj}
\end{align}
where the first term favors individually relevant images and the second term is a relevance-weighted facility-location coverage term that rewards diverse coverage of other relevant images.

Because $F_{\text{SIS}}$ is monotone submodular, the standard greedy algorithm gives a $(1-1/e)$ approximation under the cardinality constraint:
\begin{align}
A_0 &= \varnothing,\\
i_t &= \arg\max_{i\notin A_{t-1}} \Delta(i\mid A_{t-1}),\\
A_t &= A_{t-1}\cup\{i_t\},\qquad t=1,\dots,K,
\end{align}
where $\Delta(i\mid A)=F_{\text{SIS}}(A\cup\{i\})-F_{\text{SIS}}(A)$. Maintaining cached maxima
\begin{align}
m_j(A)=\max_{p\in A}\tilde d_{pj}
\end{align}
where we use the convention $m_j(\varnothing)=0$. This yields
\begin{align}
\Delta(i\mid A)
=
\lambda_{\text{rel}}\tilde r_i
+
\lambda_{\text{cov}}\sum_{j=1}^{N}\tilde r_j
\max\bigl(0,\tilde d_{ij}-m_j(A)\bigr).
\end{align}

\paragraph{Granularity.}
In experiments, we apply SIS at the entity level for MNER and at the entity-pair level for MRE, unless otherwise stated.

\paragraph{Complexity.}
Computing all $\tilde r_i$ costs $O(Nd_c)$ and all pairwise $\tilde d_{ij}$ costs $O(N^2 d_c)$. With cached $m_j(A)$, one greedy round costs $O(N^2)$ and the full selection costs $O(KN^2)$ in the worst case. 
In our multi-image benchmarks, $N$ and $K$ are small, so SIS overhead is minor relative to region-level encoding and cross-modal fusion.

\subsubsection{Optional RL selector (RES)}
\label{sec:rl}

We also study a sequential selector, RES, which formulates evidence acquisition as an MDP over the text, the current selected set, and budget signals. The policy selects one remaining image or STOP and is optimized with a cost-aware reward under CGG-based action masking. In practice, we warm-start from SIS trajectories and perform short PPO-style fine-tuning. Because SIS is more stable and directly comparable across benchmarks, all main tables use the SIS variant and RES is discussed only in the ablations.

\subsection{Energy-Inspired Joint Scoring Head}
\label{subsec:energy}

We keep the energy notation as a compact way to couple task scores, consistency regularization, and sparse vision routing, while training with standard cross-entropy losses rather than iterative energy minimization.

For MNER, over candidate spans $s$, we score span validity $\delta_s$ and type $y_s$ by
\begin{align}
E_\theta^{\mathrm{ner}}(Y^{\mathrm{ner}}\mid x,\mathcal I)
&= \sum_s \phi_{\text{span}}(\tilde h_s,\delta_s)
 + \sum_{s:\delta_s=1} \phi_{\text{type}}(\tilde h_s,y_s) \nonumber\\
&\quad + \lambda_{\text{cons}}\sum_s \eta_s\, f_{\text{cons}}(\mathrm{Consis}(s))
 + \lambda_{\text{gate}}\sum_s \eta_s.
\end{align}
where $\delta_s$ is the span-validity variable for candidate span $s$.

For MRE, given the marked entity pair $(s_h,s_t)$, we score each relation label $r$ by
\begin{align}
E_\theta^{\mathrm{re}}(r\mid x,\mathcal I)
&= \phi_{\text{rel}}(\tilde u_{(h,t)},r)
 + \lambda_{\text{cons}} \eta_{(h,t)}\, f_{\text{cons}}(\mathrm{Consis}(h,t))
 + \lambda_{\text{gate}}\eta_{(h,t)}.
\end{align}
Here $\mathrm{Consis}(h,t)$ is evaluated only when pair-level evidence is built; otherwise, we set $\mathrm{Consis}(h,t)=0$.

We use soft gates $\eta$ in the training objective and reserve hard indicators $\gamma$ for test-time routing and reporting.

At training time, these energies are treated as negative logits for standard task losses. For MNER, we enumerate candidate spans up to a maximum length $L_{\max}=10$ and keep the non-overlapping span/type assignments with the lowest energies using greedy decoding in ascending energy order. For MRE, we predict
\begin{align}
r^\star=\arg\min_r E_\theta^{\mathrm{re}}(r\mid x,\mathcal I).
\end{align}
\subsection{Complexity and deployment}
\label{sec:complexity}

Let $\mathcal{F}_T$ denote text encoding FLOPs, $\mathcal{F}_{V}^{\text{glob}}$ global image encoding FLOPs, $\mathcal{F}_{V}^{\text{reg}}(K)$ region-level FLOPs for $K$ selected images, $\mathcal{F}_{\text{fuse}}(K)$ fusion FLOPs, and $\mathcal{F}_{\text{head}}$ task-head FLOPs. Let $\mathcal{U}(x)$ denote the set of task units in sample $x$ and define the sample-level activated-unit ratio as
\begin{align}
\bar{\gamma}(x)=\frac{1}{|\mathcal{U}(x)|}\sum_{u\in\mathcal{U}(x)}\gamma(u).
\end{align}
For MRE, each instance contains one marked pair, so $\bar{\gamma}(x)=\gamma_{(h,t)}$; for MNER, $\bar{\gamma}(x)$ is the fraction of candidate spans routed to the visual branch. Because CGG uses only cached global image embeddings, unactivated units skip region extraction and multimodal fusion, giving the approximate per-sample cost
\begin{align}
\mathcal{F}_{\text{SAVER}}
\approx
\mathcal{F}_T+\mathcal{F}_{V}^{\text{glob}}+
\mathbb{E}_{x}\!\left[\bar{\gamma}(x)\right]
\bigl(\mathcal{F}_{V}^{\text{reg}}(K)+\mathcal{F}_{\text{fuse}}(K)\bigr)
+\mathcal{F}_{\text{head}}.
\end{align}
This expression is an approximation at the sample level; in practice, selected image and region features can be cached and shared across activated units within the same post.

In practice, cost grows roughly with the selected budget $K$, while cached global embeddings keep the no-activation path cheap. SAVER also remains interpretable because it returns an explicit selected image subset, Set Transformer attention weights, and consistency scores for activated units.

\section{Experiments and Analysis}
\label{sec:exp}

We introduce datasets, metrics, and baselines, then report overall performance, selectivity, calibration, efficiency, and ablation results, together with a brief sensitivity discussion and compact observations on the RES extension.
Unless noted, the text encoder is \TextEncName~\cite{warner_smarter_2024} and the vision encoder is \VisEncName~\cite{tschannen_siglip_2025}. Throughout Sec.~\ref{sec:exp}, \textbf{SAVER} refers to the default SIS-based variant, and \textbf{SAVER (full)} denotes the complete \textbf{CGG+SIS+Joint Scoring} model.

\subsection{Datasets and evaluation protocol}
\label{subsec:datasets}

We evaluate on two RE benchmarks (MNRE~\cite{zheng_mnre_2021} and MRE-MI~\cite{huang_mre-mi_2025}) and four MNER benchmarks (Twitter-2015 / Twitter-2017~\cite{zhang_adaptive_2018,lu_visual_2018}, MNER-MI~\cite{huang_mner-mi_2024}, and MNER-MI-Plus~\cite{huang_mner-mi_2024}) using the official splits. MNRE contains 12{,}247 / 1{,}624 / 1{,}614 instances with one image per sample, MRE-MI contains 13{,}504 / 4{,}500 / 4{,}500 instances with 2.80 images on average, Twitter-2017 contains 3{,}373 / 723 / 723 posts, and MNER-MI-Plus contains 10{,}229 / 1{,}583 / 1{,}583 posts with 2.15 images on average. For completeness, Twitter-2015 contains 4{,}000 / 1{,}000 / 3{,}257 samples and MNER-MI contains 6{,}856 / 860 / 860.

\noindent\textbf{MNRE (single-image MRE).} We use the refined MNRE (v2) split~\cite{zheng_mnre_2021}, which is widely adopted in prior work, and follow its established evaluation protocol.

\noindent\textbf{MRE-MI (multi-image RE).} MRE-MI is a multi-image human-annotated MRE dataset containing both single- and multi-image instances, with strong baselines and official evaluation scripts~\cite{huang_mre-mi_2025}. We use its official split to evaluate relevance--diversity selection and selective vision usage.

\noindent\textbf{Twitter-2015 / Twitter-2017 and MNER-MI / MNER-MI-Plus.} These are standard single- and multi-image MNER benchmarks~\cite{zhang_adaptive_2018,lu_visual_2018,huang_mner-mi_2024}. We follow the official splits and common evaluation protocols, and use them to test whether selective visual evidence benefits both single-image and multi-image settings.

For CGG calibration, we use the official dev split for model selection and reserve $10\%$ of the training split for calibration; $\tau$ is calibrated separately for each dataset at $\alpha=0.10$ and $1-\delta=0.95$.

\subsection{Evaluation metrics}
\label{subsec:metrics}

For MRE, we report micro-F1 on relation classification for marked entity pairs, together with precision and recall. For MNER, we report strict entity-level F1 on span boundary and type. To evaluate selectivity, we compute risk--activation-coverage curves at the task unit (candidate-span level for MNER and marked-pair level for MRE), AURC, and ActCov@0.10. For SAVER, activation coverage is induced by thresholding CGG scores; for baselines without CGG, activation coverage is induced by thresholding model confidence to obtain comparable curves. For methods without an explicit routing gate, ActCov@0.10 is reported only for comparing risk--activation-coverage trade-offs and should not be interpreted as an actual vision-routing rate. We additionally report per-sample FLOPs, end-to-end P90 latency, and activation coverage (reported as ActCov@0.10 in the main tables).

\subsection{Baselines}
\label{subsec:baselines}

We compare against three groups of baselines. For MNER, text-only baselines include BERT-CRF and RoBERTa-CRF, built on BERT~\cite{devlin_bert_2019}, RoBERTa~\cite{liu_roberta_2019}, and a CRF decoder~\cite{lafferty_conditional_2001}, as well as \TextEncName~\cite{warner_smarter_2024} and \TextEncAlt~\cite{he_debertav3_2023}. For MRE, text-only baselines include \TextEncName\ and \TextEncAlt. We further compare against always-on multimodal models (HVPNeT~\cite{chen_good_2022}, RSRNeT~\cite{wang_rsrnet_2024}, and retrieval-augmented MRE~\cite{wang_named_2022}) and multi-image baselines (All-Images Attn., Top-$K$ by relevance, GLRA~\cite{huang_mre-mi_2025} on MRE-MI, and \textbf{GLRA-adapt}, our adaptation of GLRA to MNER-MI and MNER-MI-Plus, on the multi-image MNER benchmarks). Unless noted, multimodal baselines are reimplemented with the same default text and vision backbones as SAVER for fair comparison. We also report an RL-based variant, SAVER-RL, in addition to the default SIS-based model.

\subsection{Results and Analysis}
\label{subsec:results}

\subsubsection{Overall performance across RE and NER benchmarks}
\label{sec:full-results}

\noindent\textbf{Main benchmark: MRE-MI.}
\begin{table}[tb]
\centering
\scriptsize
\setlength{\tabcolsep}{2.5pt}
\caption{MRE-MI results. SAVER (full) denotes the complete CGG+SIS+Joint Scoring model; SAVER w/o J.Score replaces the joint-scoring head with a standard task head; SAVER (CGG+SIS) removes the consistency and sparsity terms.}

\label{tab:main_mremi}
\begin{tabular}{lccccccc}
\toprule
Method & P\up & R\up & F1\up & AURC\down & ActCov@0.10\up & FLOPs (G/sample)\down & P90 (ms)\down \\
\midrule
\multicolumn{8}{l}{\textit{Text-only}} \\
ModernBERT-only & 82.37 & 79.84 & 81.09 & 0.147 & 0.68 & 13 & 17 \\
DeBERTa-v3-only & 81.53 & 79.48 & 80.49 & 0.153 & 0.66 & 18 & 27 \\
\midrule
\multicolumn{8}{l}{\textit{Multimodal}} \\
HVPNeT & 73.87 & 76.82 & 75.32 & 0.168 & 0.63 & 66 & 99 \\
RSRNeT & 84.78 & 83.06 & 83.89 & 0.129 & 0.74 & 60 & 90 \\
All-Images Attn. & 83.47 & 82.18 & 82.82 & 0.142 & 0.72 & 62 & 93 \\
Top-$K$ by relevance & 85.31 & 83.62 & 84.45 & 0.119 & 0.77 & 51 & 77 \\
GLRA & 85.23 & 83.81 & 84.51 & 0.117 & 0.78 & 56 & 84 \\
Retrieval-Aug. & 84.27 & 82.86 & 83.56 & 0.124 & 0.75 & 55 & 82 \\
\midrule
\multicolumn{8}{l}{\textit{Ours/Abl.}} \\
\textbf{SAVER (full)} & 85.93 & 84.57 & \textbf{85.24} & \textbf{0.104} & \textbf{0.82} & \textbf{36} & \textbf{54} \\
SAVER w/o CGG & 84.46 & 83.18 & 83.81 & 0.124 & 0.76 & 51 & 77 \\
SAVER w/o SIS & 84.13 & 82.74 & 83.43 & 0.136 & 0.74 & 62 & 93 \\
SAVER w/o J.Score & 85.28 & 84.12 & 84.70 & 0.111 & 0.80 & 37 & 56 \\
SAVER (CGG+SIS) & 85.14 & 84.33 & 84.73 & 0.107 & 0.81 & 35 & 53 \\
\bottomrule
\end{tabular}
\end{table}

SAVER achieves the best F1 on the core multi-image RE benchmark while also improving ActCov@0.10 and reducing FLOPs/P90 latency. The results are consistent with the design intuition that CGG suppresses unnecessary vision calls and SIS retains complementary evidence. The ablations in the same table show that CGG, SIS, and the energy-inspired joint scoring head each contribute independently.

\noindent\textbf{Additional benchmarks.}

\begin{table}[tb]
\centering
\caption{Additional benchmarks (F1). Strong baseline denotes the best included non-SAVER baseline for each dataset.}
\label{tab:additional_all}
\begin{tabular}{lccc}
\toprule
Dataset & Strong baseline & SAVER & Gain \\
\midrule
MNRE & 83.9 (RSRNeT) & \textbf{84.7} & +0.8 \\
Twitter-2015 & 76.5 (RSRNeT) & \textbf{77.0} & +0.5 \\
Twitter-2017 & 87.9 (RSRNeT) & \textbf{88.0} & +0.1 \\
MNER-MI & 76.9 (GLRA-adapt) & \textbf{77.3} & +0.4 \\
MNER-MI-Plus & 83.1 (GLRA-adapt) & \textbf{83.7} & +0.6 \\
\bottomrule
\end{tabular}
\end{table}

Across the remaining single- and multi-image benchmarks, SAVER consistently outperforms the best included non-SAVER baseline. The gains are smaller on single-image settings and clearer again on multi-image settings, which matches the fact that selective visual usage matters most when images are noisy, weakly aligned, or redundant.

\subsubsection{Selectivity, calibration, and efficiency}
\label{sec:selectivity-efficiency}

\begin{table}[tb]
\centering
\caption{CGG in-domain calibration at $\alpha=0.10$ ($1-\delta=0.95$).}
\label{tab:cgg_crc_in}
\begin{tabular}{lccc}
\toprule
Dataset & Act. Coverage & Emp.\ Error & CP Upper \\
\midrule
MRE-MI & 0.82 & 0.087 & 0.097 \\
MNRE & 0.82 & 0.083 & 0.098 \\
Twitter-2017 & 0.80 & 0.077 & 0.099 \\
MNER-MI-Plus & 0.81 & 0.084 & 0.099 \\
\bottomrule
\end{tabular}
\end{table}

\begin{figure}[tb]
  \centering
  \includegraphics[width=\linewidth]{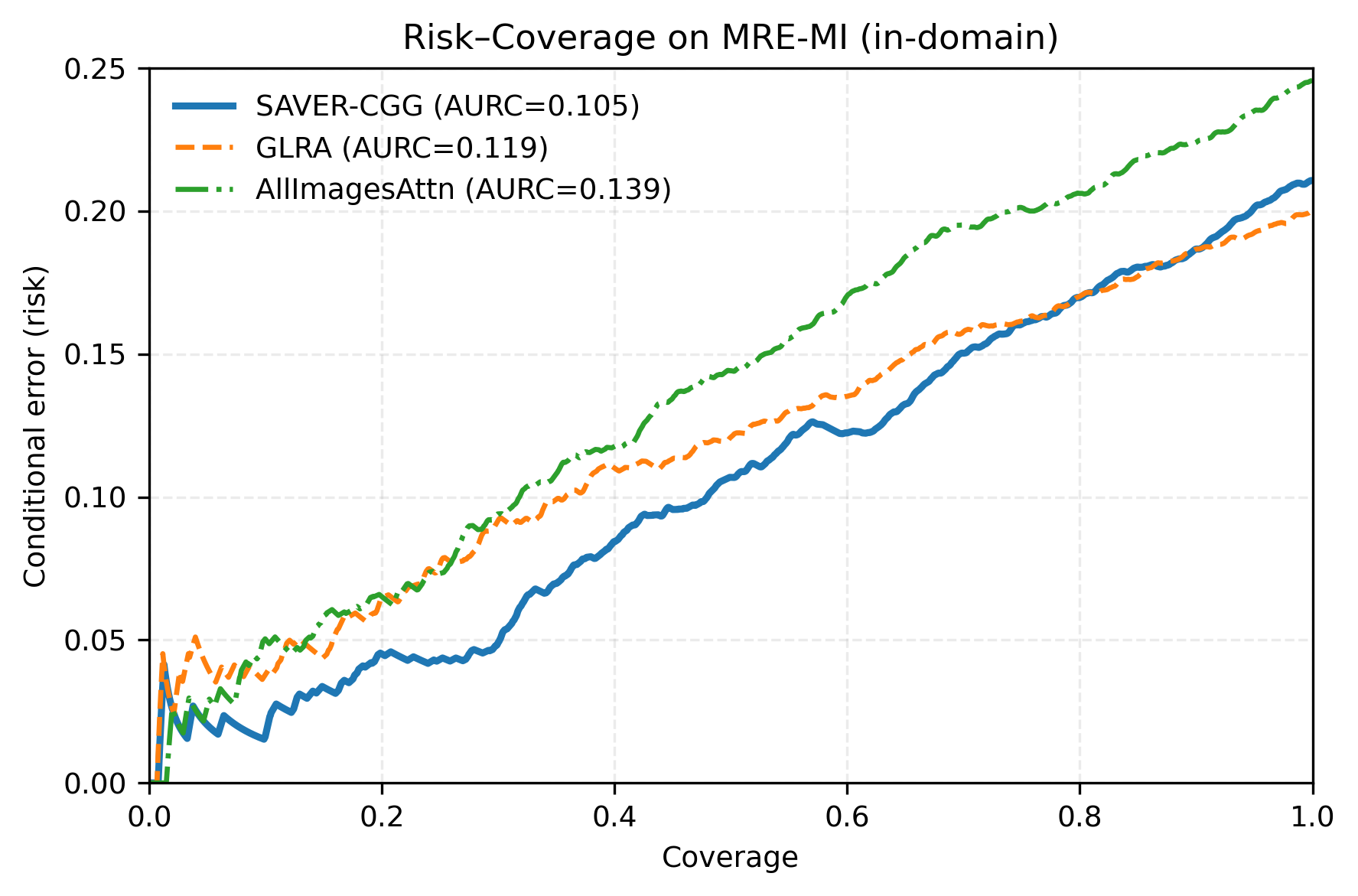}
  \caption{Risk--activation-coverage curves on MRE-MI. SAVER uses split-calibrated CGG with $\alpha=0.10$; baselines use confidence thresholding. AURC is shown in the legend.}
  \label{fig:rc_cgg}
\end{figure}

In-domain calibration keeps the activated-subset error close to the target level. Fig.~\ref{fig:rc_cgg} shows that SAVER still dominates confidence-based gating baselines along the risk--activation-coverage curve. 
Efficiency is already summarized in Table~\ref{tab:main_mremi}: on MRE-MI, SAVER reduces cost from 60G/90\,ms (RSRNeT) to 36G/54\,ms; at the $\alpha=0.10$ operating point, the activation coverage is 0.82, while preserving the best overall accuracy/selectivity trade-off.

\subsubsection{Ablations and sensitivity analyses}
\label{sec:ablations}

Table~\ref{tab:main_mremi} shows that CGG, SIS, and the energy-inspired joint scoring head each contribute independently. Removing CGG increases compute and weakens the overall risk--efficiency trade-off; removing SIS weakens the multi-image gain; and removing the energy coupling slightly lowers both accuracy and selectivity. We use $\alpha=0.10$ in the main experiments as a balanced operating point between risk, activation coverage, and efficiency. Across the tested settings, smaller $\alpha$ makes the gate more conservative, while performance increases with $K$ up to a small budget and then saturates; this is consistent with the goal of selecting only a few complementary images.

For the optional RES extension, removing reward shaping slows convergence and degrades selectivity; offline-only policies lag behind short on-policy fine-tuning; and disabling the CGG action mask increases both risk and cost. These observations are consistent with the role of CGG as a safe candidate filter and SIS trajectories as a strong warm start. We therefore keep SIS as the default selector in the main comparison tables and present RES as a compact exploratory extension rather than a replacement for SIS.

\FloatBarrier
\section{Conclusion}
\label{sec:conclusion}

We presented SAVER, a selective multimodal IE framework that treats vision as optional evidence. A calibrated gate decides whether to activate vision, and a compact relevance--diversity selector acquires a small evidence set that is fused with text by a joint scoring head. Across MRE and MNER benchmarks, SAVER improves F1 while reducing AURC, FLOPs, and P90 latency, with larger gains in multi-image settings. The framework is modular and can be paired with alternative encoders or evidence selectors.

\bibliographystyle{splncs04}
\bibliography{260325}
\end{document}